\documentclass[sn-basic, Numbered,pdflatex]{sn-jnl}

\usepackage{amsmath}     
\usepackage{amssymb}
\usepackage{graphicx}
\usepackage{multirow}    
\usepackage{booktabs}    
\usepackage{url}

\graphicspath{{figures/}}

\begin{document}

\title[Spatial Attention Supervision for Defect Localization]{Spatial Attention Supervision for Defect Localization: Exploiting Ground-Truth Masks as Training Signal in Diffusion-Augmented Defect Detection}

\author*{\fnm{Sajjad} \sur{Rezvani Boroujeni}}\email{saj@actualreality.tech}
\author{\fnm{Muskan} \sur{Saraf}}\email{muskan@actualreality.tech}
\author{\fnm{Gnana Tulasi} \sur{Makineni}}\email{gnana@actualreality.tech}
\author{\fnm{Tom} \sur{Bush}}\email{tom@actualreality.tech}
\author{\fnm{Hossein} \sur{Abedi}}\email{hossein@actualreality.tech}

\affil{\orgdiv{Data Science Department}, \orgname{Actual Reality Technologies}, \orgaddress{\state{OH}, \country{USA}}}

\abstract{Ground-truth defect masks in industrial inspection datasets are typically reserved for evaluation. This paper repurposes them as \emph{spatial supervision signals} during the training of classification networks, teaching a model not just what to predict but where to look. The proposed method adds an activation-based attention alignment loss that steers convolutional feature maps toward defect regions, within a mixed-supervision formulation that also accommodates samples without masks, such as diffusion-generated images. Combined with DDPM-based data augmentation, this produces a training regime in which synthetic images contribute quantity and the masks contribute spatial precision. The approach is evaluated across 85 trained models (four CNN backbones under a $2 \times 2$ data/training factorial design over five seeds, plus a Swin-V2-T transformer baseline over five seeds) on the MVTec-AD bottle benchmark. All localization is evaluated on \emph{held-out} defect images excluded from classifier gradient updates, with metrics aggregated as mean $\pm$ standard deviation across five seeds. The main findings are: (1)~on these held-out images, attention-guided training improves activation-based localization (Pixel-AUROC) by +18.0\% for EfficientNetB0 in the augmented setting ($p = 0.005$, Cohen's $d = 2.6$) and by +18.7\% for ResNet50 ($p = 0.008$), with gains reaching significance in four of the eight CNN settings (all four among the three backbones that respond to the loss; uncorrected for multiple comparisons) and no statistically significant change in classification; (2)~for EfficientNetB0 a data $\times$ training-mode interaction test is significant ($p = 0.002$), consistent with a super-additive effect: combining the two improves localization by +13.6\% over the real-data, standard-training baseline, beyond the sum of their individual effects (+1.6\%; augmentation alone is slightly negative and attention alone gives +5.4\%); (3)~architectures with weaker spatial representations benefit most, whereas ConvNeXt-T shows no effect, which appears to stem from its depthwise-convolution activations yielding spatially uninformative channel-mean maps; (4)~unsupervised PatchCore remains the strongest localizer (Pixel-AUROC=0.983) and serves as a strong reference that contextualizes the supervised gains. Overall, these results show that existing evaluation masks can act as practical training signals that measurably and reproducibly improve where defect classifiers attend.}


\maketitle

\section{Introduction}\label{sec:introduction}

Automated visual inspection using deep learning has become essential in glass container manufacturing, where defects such as cracks, bubbles, and contamination compromise product integrity and safety~\cite{grand_view_2022, hutten2024deep, makineni2026highvariation}. While binary classification (labeling images as defective or non-defective) has seen substantial progress, industrial deployment increasingly demands \emph{localization}: identifying not just whether a defect exists, but where it is located. Localization supports root-cause analysis, process improvement, and operator trust in automated systems.

Prior work~\cite{rezvani2025enhancing} showed that Denoising Diffusion Probabilistic Model (DDPM)-generated synthetic defective images improve CNN-based classification on the MVTec-AD bottle dataset, raising ResNet50V2 accuracy from 78\% to 93\%. That study, like the broader literature on diffusion-augmented defect detection, reports only image-level metrics (accuracy, F1, AUROC), which leaves a basic question open: \emph{when the model makes its decision, does it actually look at the defect?}

Ground-truth defect masks exist in standard benchmarks like MVTec-AD and are routinely used for \emph{evaluation} of localization methods. However, these masks represent untapped potential: they could serve as \emph{spatial supervision signals during training} of classification networks. While segmentation models (e.g., U-Net) use masks as direct output targets, little prior work uses them as \emph{auxiliary attention guidance} for classifiers, particularly in a mixed-supervision setting where some training samples have masks and others do not.

This paper introduces spatial attention supervision for defect classification networks. The approach adds an activation-based attention alignment loss during training that encourages convolutional feature maps to activate on actual defect regions, as defined by ground-truth masks. The method operates in a \emph{mixed-supervision} framework: it uses masks when available and falls back to classification-only loss when they are not. Combined with DDPM augmentation, this creates a dual-path strategy where synthetic data provides \emph{quantity} and mask supervision provides \emph{spatial precision}.

The main contributions are:
\begin{itemize}
    \item \textbf{C1:} A lightweight activation-alignment approach that uses ground-truth defect masks as an auxiliary spatial-supervision signal (not a direct output target) for classification networks, studied in the specific setting of \emph{partial}-mask supervision combined with maskless DDPM-generated defective samples within a mixed-supervision formulation.
    \item \textbf{C2:} A $2 \times 2$ study of DDPM augmentation and attention guidance across four CNN architectures, with Swin-V2-T as a transformer baseline, revealing an architecture-dependent interaction: a significant super-additive interaction for EfficientNetB0 ($p = 0.002$), no significant interaction for MobileNetV2 (approximately additive), and a non-significant sub-additive trend for ResNet50.
    \item \textbf{C3:} A multi-architecture benchmark (85 models, 5 backbones, 5 seeds) with held-out classification evaluation and dual localization assessment (activation-based and Grad-CAM), revealing architecture-dependent responses to attention supervision.
    \item \textbf{C4:} An evaluation protocol in which localization is measured only on held-out validation images excluded from classifier gradient updates, aggregated across five seeds with paired significance tests and effect sizes, in contrast to prior work that reports single-run localization on training images.
\end{itemize}

\section{Related Work}\label{sec:related_work}

\subsection{Diffusion Models for Defect Synthesis}

DDPMs~\cite{ho2020ddpm} and their variants~\cite{nichol2021improved, dhariwal2021diffusion} have been applied to industrial defect synthesis with increasing success. Lei\~{n}ena et al.~\cite{leinena2024sensors} used Stable Diffusion for steel surface defect segmentation. Li et al.~\cite{zhang2025sensors} proposed few-shot defect synthesis with mask-guided diffusion. AnomalyDiffusion~\cite{hu2024anomalydiffusion} achieved 99.1\% pixel-AUROC on MVTec via disentangled embeddings. DefectFill~\cite{defectfill2025} introduced mask-conditioned inpainting diffusion for defect generation. Rezvani Boroujeni et al.~\cite{rezvani2025enhancing} applied an unconditional DDPM to MVTec-AD bottles, generating 60 synthetic defective images that improved CNN classification. In every case, however, the masks (where used) drive the \emph{generator}; none of the approaches reviewed here feed masks back as \emph{training supervision} for a downstream classifier.

\subsection{Explainability and Attention in Defect Detection}

Grad-CAM~\cite{selvaraju2017gradcam} produces saliency maps highlighting regions influencing model decisions. Schlosser et al.~\cite{schlosser2024glass} applied Grad-CAM to glass bottle print defect detection. Kasem et al.~\cite{attention_hybrid_2025} integrate CBAM attention modules into a YOLOv11/EfficientNet hybrid for defect classification. Region-Aware CAM~\cite{regioncam2025} proposed filtering-guided backpropagation for weakly-supervised defect segmentation. These works either use attention as an \emph{architectural component} (CBAM modules) or as a \emph{post-hoc visualization} (Grad-CAM), but none of the reviewed works use spatial attention as a \emph{training loss} for classification networks.

\subsection{Attention Supervision in Deep Learning}

Ross et al.~\cite{ross2017right} proposed penalizing input gradients outside annotated regions (``Right for the Right Reasons''), applied to sentiment analysis and medical imaging. Zhou et al.~\cite{zhou2016cam} showed that class activation maps (CAMs) from convolutional features provide spatial localization. Self-produced guidance~\cite{spg2018} uses masks as auxiliary pixel-level supervision for weakly-supervised object localization. More directly, the Guided Attention Inference Network~\cite{gain2018} makes attention maps an explicit target of end-to-end training, including a variant that supervises attention with a small subset of pixel-level masks, and the Attention Branch Network~\cite{abn2019} learns an attention branch that both improves accuracy and yields a visual explanation. This line of work establishes that attention maps can be supervised, including with partial masks, but on natural-image classification and localization benchmarks. To the extent reviewed here, it has not been applied to industrial defect detection, nor studied in combination with diffusion-generated maskless samples or across a panel of modern backbones. Those specific combinations, rather than attention supervision per se, are the focus of this paper.

\subsection{Unsupervised Anomaly Localization}

PatchCore~\cite{roth2022patchcore} memorizes a coreset of normal patch features and scores test patches by nearest-neighbor distance. EfficientAD~\cite{batzner2024efficientad} and FastFlow~\cite{yu2021fastflow} report 100\% image-level AUROC on the bottle category. These methods produce pixel-level anomaly maps as a natural byproduct. The MVTec AD~2 dataset~\cite{mvtecad2_2026} was introduced specifically because existing benchmarks have saturated. PatchCore is included as an unsupervised localization baseline to contextualize the supervised results.

\subsection{Positioning of This Work}

Table~\ref{tab:positioning} summarizes how the proposed approach relates to prior work. The key distinction is the use of ground-truth masks as an \emph{auxiliary attention training signal} for a classifier (output: class label), rather than as a direct output target (segmentation) or post-hoc visualization (Grad-CAM).

\begin{table}[!t]
\caption{Positioning relative to prior work. Among the approaches reviewed, the proposed method is distinctive in combining auxiliary mask supervision with diffusion augmentation in a mixed-supervision setting.}\label{tab:positioning}
\begin{tabular}{@{}p{3.4cm}p{3.4cm}p{4.8cm}@{}}
\toprule
\textbf{Approach} & \textbf{Use of Masks} & \textbf{Difference from Proposed} \\
\midrule
Segmentation (U-Net, DeepLab) & Direct output target & Predicts mask; typically trained with dense mask supervision \\
Ross et al.~\cite{ross2017right} & Gradient penalty & Input gradients (not activations); not industrial \\
Kasem et al.~\cite{attention_hybrid_2025} & None (architectural) & CBAM modules in architecture, not a training loss \\
Region-Aware CAM~\cite{regioncam2025} & Evaluation only & Weakly-supervised; no mask-guided training \\
PatchCore~\cite{roth2022patchcore} & Evaluation only & Unsupervised; masks unused during training \\
Prior work~\cite{rezvani2025enhancing} & Evaluation only & DDPM augmentation for classification; no localization \\
\midrule
\textbf{This work} & \textbf{Auxiliary attention loss} & \textbf{Classifier + mixed supervision + DDPM} \\
\bottomrule
\end{tabular}
\end{table}

\section{Methodology}\label{sec:methodology}

\begin{figure}[!t]
    \centering
    \includegraphics[width=\textwidth]{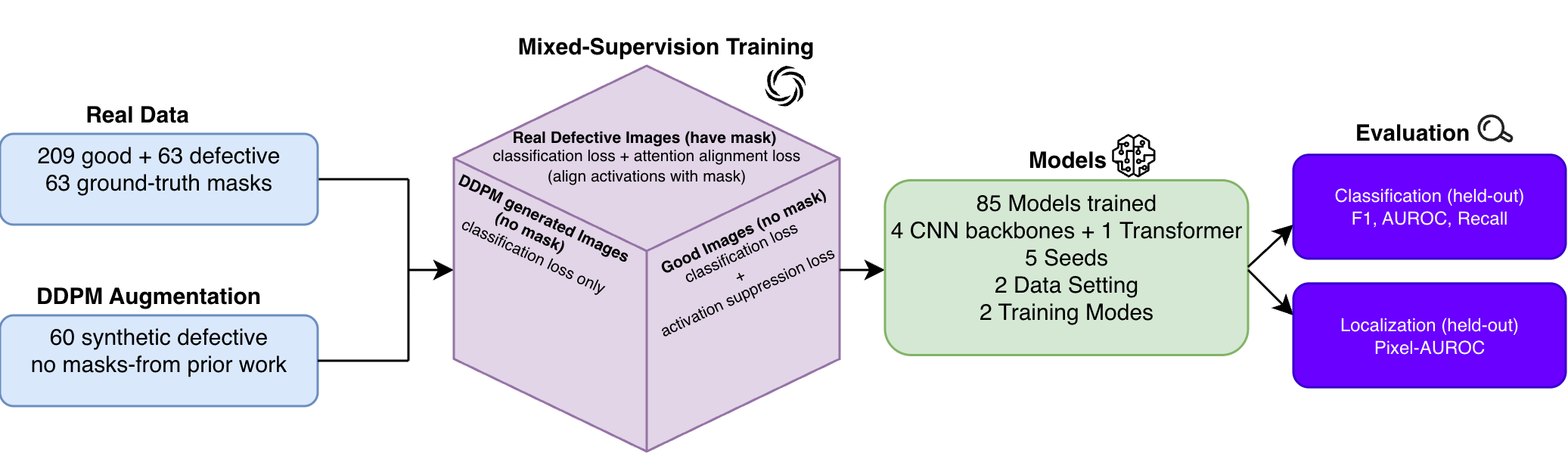}
    \caption{Pipeline overview. Data: MVTec-AD bottle with optional DDPM augmentation. Training: 85 models total, four CNN backbones under a $2 \times 2$ factorial (data setting $\times$ training mode) across five seeds, plus Swin-V2-T baseline across five seeds. Evaluation: held-out classification metrics and held-out localization via Pixel-AUROC.}
    \label{fig:pipeline}
\end{figure}

Figure~\ref{fig:pipeline} provides an overview of the complete pipeline.

\subsection{Dataset}

This work uses the bottle category of MVTec-AD~\cite{bergmann2019mvtec} (CC BY-NC-SA 4.0), identical to the prior study~\cite{rezvani2025enhancing}. The dataset comprises 209 non-defective training images, 20 non-defective test images, and 63 defective test images across three types: broken large (20), broken small (22), and contamination (21). Each defective image has a corresponding pixel-level binary mask. Following~\cite{rezvani2025enhancing}, 60 DDPM-generated synthetic defective images augment the training set in the ``AugmentedData'' configuration. Note that the DDPM is unconditional: it produces defective images but \emph{not} corresponding masks. These synthetic images were generated once in the prior study~\cite{rezvani2025enhancing} from the real defective set, before the per-seed held-out protocol used here was defined; the augmented-data setting may therefore involve indirect exposure to defective images that are later held out for evaluation, a limitation discussed in Section~\ref{sec:limitations}.

Two training configurations:
\begin{itemize}
    \item \textbf{RealData:} 209 good + 63 real defective (masks available for all 63)
    \item \textbf{AugmentedData:} 209 good + 63 real defective + 60 DDPM-generated (masks available for 63 real only)
\end{itemize}

Classification metrics are evaluated on a held-out set: the 20 good test images (never in training) plus the defective images held out during the stratified train/validation split (varying per seed). This avoids train-test overlap for classification evaluation.

\subsection{Backbone Architectures}

Five architectures are evaluated, spanning lightweight to modern CNNs and a vision transformer (Table~\ref{tab:backbones}).

\begin{table}[!t]
\caption{Backbone Architectures.}\label{tab:backbones}
\begin{tabular}{@{}lllrl@{}}
\toprule
\textbf{Backbone} & \textbf{Type} & \textbf{Year} & \textbf{Params} & \textbf{Input Size} \\
\midrule
EfficientNetB0~\cite{tan2019efficientnet} & CNN & 2019 & 5.3M & 224$\times$224 \\
MobileNetV2~\cite{sandler2018mobilenetv2} & CNN & 2018 & 3.5M & 224$\times$224 \\
ResNet50~\cite{he2016resnet} & CNN & 2016 & 25.6M & 224$\times$224 \\
ConvNeXt-T~\cite{liu2022convnext} & CNN & 2022 & 28.6M & 224$\times$224 \\
Swin-V2-T~\cite{liu2022swinv2} & Transformer & 2022 & 28M & 256$\times$256 \\
\bottomrule
\end{tabular}
\end{table}

EfficientNetB0 and MobileNetV2 were used in the prior study~\cite{rezvani2025enhancing}. ResNet50 is the standard benchmark CNN. ConvNeXt-T represents modern CNN design inspired by transformers. Swin-V2-T serves as the transformer baseline (standard training only, as attention supervision via convolutional activation maps is not applicable to transformers). All models use ImageNet-1K pretrained weights from the \texttt{timm} library~\cite{timm}.

\subsection{Training Protocol}

All models use a Linear Probe then Fine-Tune (LP-FT) strategy:
\begin{enumerate}
    \item \textbf{Linear probe} (5 epochs): freeze backbone, train classification head only.
    \item \textbf{Fine-tune} (up to 25 epochs): unfreeze all layers at $0.1\times$ learning rate.
\end{enumerate}

Hyperparameters: AdamW optimizer, $\text{lr} = 10^{-4}$, weight decay $10^{-4}$, cosine annealing, batch size 16, early stopping with patience 5, gradient clipping at norm 1.0. Standard augmentation (random flips, rotation $\pm 20^{\circ}$, zoom $\pm 20\%$, contrast $\pm 20\%$). Class weights follow~\cite{rezvani2025enhancing}: $w_0 = N / (N_{\text{good}} \times 1.2)$ and $w_1 = N / (N_{\text{def}} \times 2.1)$, with $N$ the training-set size. This formula is kept unchanged for consistency with prior work. Because it scales with class counts, the effective up-weighting of the defective class differs between settings: in RealData ($N_{\text{good}}{=}209$, $N_{\text{def}}{=}63$) defectives receive $\approx 1.9\times$ the weight of good samples, whereas in AugmentedData the 60 synthetic defectives enlarge $N_{\text{def}}$ to 123 and the ratio falls to $\approx 0.97\times$. This lighter re-weighting is reasonable, since the synthetic defectives already rebalance the training distribution. It also has a bounded effect on the conclusions: the cleanest evidence for attention guidance is the standard-versus-attention-guided comparison \emph{within} a data setting, where both modes share identical class weights and the difference cancels. The cross-setting factorial decomposition (Section~\ref{sec:complementarity}) does compare RealData and AugmentedData and therefore inherits this difference along with the effect of the added data; those interaction terms are treated as indicative rather than definitive, with the caveat noted in Section~\ref{sec:limitations}. Decision threshold: 0.4 (chosen a priori, not tuned on the evaluation data). Each configuration is repeated with 5 seeds (42, 123, 456, 789, 1024); see Section~\ref{sec:experimental_setup} for the full factorial design.

\subsection{Spatial Attention Supervision}
\label{sec:attention_supervision}

The core methodological contribution is a mixed-supervision training loss that uses ground-truth masks to guide the spatial activations of the classification network. Unlike segmentation, where the mask is the \emph{output target}, here the mask serves as an \emph{auxiliary spatial supervision signal}; the model's output remains a binary class label.

\subsubsection{Activation-Based Spatial Attention}

During the forward pass, the activations $\mathbf{A} \in \mathbb{R}^{C \times H' \times W'}$ are captured from the last convolutional layer via a forward hook. This last convolutional layer is a $1 \times 1$ pointwise (channel-mixing) convolution for EfficientNetB0, MobileNetV2, and ResNet50, but a $7 \times 7$ depthwise convolution for ConvNeXt-T; this distinction is revisited in Section~\ref{sec:convnext}. The spatial attention map is computed as the channel-wise mean followed by normalization:
\begin{equation}
    \hat{A}(h, w) = \frac{\text{ReLU}\left(\frac{1}{C}\sum_{c=1}^{C} A_{c,h,w}\right) - \min}{\max - \min + \epsilon}
\end{equation}
where $\epsilon = 10^{-8}$ prevents division by zero. This produces a normalized spatial attention map $\hat{A} \in [0,1]^{H' \times W'}$, resized to match the ground-truth mask resolution via bilinear interpolation.

\subsubsection{Mixed-Supervision Loss}

The total loss for attention-guided training is:
\begin{equation}
    \mathcal{L} = \mathcal{L}_{\text{cls}} + \lambda(t) \cdot \mathcal{L}_{\text{attn}}
\end{equation}

where $\mathcal{L}_{\text{cls}}$ is the weighted cross-entropy classification loss (applied to all samples) and $\mathcal{L}_{\text{attn}}$ is the spatial attention loss (applied selectively). The attention weight $\lambda(t)$ follows a warmup schedule:
\begin{equation}
    \lambda(t) = \begin{cases}
        0 & t < t_{\text{freeze}} \\
        \lambda_{\max} \cdot \frac{t - t_{\text{freeze}}}{t_{\text{warmup}}} & t_{\text{freeze}} \leq t < t_{\text{freeze}} + t_{\text{warmup}} \\
        \lambda_{\max} & t \geq t_{\text{freeze}} + t_{\text{warmup}}
    \end{cases}
\end{equation}

with $t_{\text{freeze}} = 5$ (no attention loss during the linear probe phase, when the backbone is frozen and activations are not yet meaningful), $t_{\text{warmup}} = 3$ epochs, and $\lambda_{\max} = 0.5$. These weights ($\lambda_{\max}=0.5$, the $0.5$ cosine term in Equation~\ref{eq:align}, and the $0.3$ suppression weight in Equation~\ref{eq:suppress}) were fixed a priori rather than tuned, to avoid selection on the small evaluation set.

The attention loss $\mathcal{L}_{\text{attn}}$ routes differently based on sample type:

\textbf{Real defective images (have mask):}
\begin{equation}
    \mathcal{L}_{\text{align}} = \text{MSE}(\hat{A}, M) + 0.5 \cdot (1 - \cos(\hat{A}, M))
    \label{eq:align}
\end{equation}
where $M$ is the ground-truth mask resized to match $\hat{A}$, and $\cos(\cdot, \cdot)$ is cosine similarity computed over flattened spatial dimensions. This encourages the model to activate on defect regions.

\textbf{DDPM-generated images (no mask):} $\mathcal{L}_{\text{attn}} = 0$. These samples contribute only to $\mathcal{L}_{\text{cls}}$, as the defect location is unknown in synthetic images.

\textbf{Good images (no defect):}
\begin{equation}
    \mathcal{L}_{\text{suppress}} = 0.3 \cdot \text{mean}(\hat{A})
    \label{eq:suppress}
\end{equation}
This penalizes strong activations on defect-free images, as there is nothing to localize.

\subsection{Localization Evaluation}
\label{sec:methodology_localization}

Two localization methods are used for evaluation:

\textbf{Activation-based} (matches training): The same spatial attention map used during training (Equation 1) is computed at test time and compared against ground-truth masks. This provides a fair assessment of whether the attention supervision affected the model's internal spatial representations.

\textbf{Grad-CAM}~\cite{selvaraju2017gradcam} (post-hoc): Standard Grad-CAM from the \texttt{captum} library~\cite{captum} provides a gradient-weighted class-discriminative heatmap. This assesses whether improvements in activation-based localization transfer to the standard explainability method.

For Swin-V2-T, occlusion sensitivity~\cite{zeiler2014visualizing} ($32 \times 32$ patches, stride 16) is used instead.

\textbf{PatchCore}~\cite{roth2022patchcore} (unsupervised baseline): Trained on the 209 good images only using ResNet-50 layer-3 features with a 5,000-patch coreset. This provides a strong localization reference from a fundamentally different paradigm. Because PatchCore never trains on defective images, it has no train/test overlap and is evaluated on all 63 defective masks.

\textbf{Held-out evaluation.} Localization quality is reported as Pixel-AUROC, the standard threshold-free localization metric for the MVTec-AD benchmark. Importantly, for the supervised models, localization is evaluated \emph{only on held-out validation images that were excluded from classifier gradient updates}. For each seed, the same stratified 15\% validation split used for held-out classification (Section~\ref{sec:experimental_setup}) defines these images; localization is computed on those images alone, which never receive gradient updates or mask supervision. This validation split is also used for early stopping and best-checkpoint selection (by validation loss), so it constitutes a held-out validation set rather than a fully untouched test set. Two properties bound the effect of this on the central result. First, checkpoints are selected by validation \emph{classification} loss, never by any localization metric, so the localization scores reported here are not optimized on the held-out images. Second, the identical protocol is applied to the standard and attention-guided modes, so it cannot by itself create a localization difference between them; accordingly, the within-setting standard-versus-attention-guided contrast (Section~\ref{sec:experimental_setup}) is the primary evidence in this paper. The 15\% split is drawn over the positive pool of each setting, which differs in size: RealData has 63 defective images (9 held out per seed), whereas AugmentedData has 123 positives (63 real plus 60 synthetic) and therefore holds out 18 per seed. Because the synthetic images carry no masks, only the \emph{real} held-out defectives can be scored for localization (roughly 11 to 12 per seed in AugmentedData). Aggregating across the five seeds thus yields 45 (RealData) and 57 (AugmentedData) held-out localization evaluations per configuration. Pixel-AUROC is reported as the mean $\pm$ standard deviation of the five per-seed means, matching the protocol used for classification.

\section{Experimental Setup}
\label{sec:experimental_setup}

\subsection{Factorial Design}

A $2 \times 2$ factorial design crosses two factors:
\begin{itemize}
    \item \textbf{Data augmentation:} RealData (272 images) vs.\ AugmentedData (+60 DDPM, 332 images)
    \item \textbf{Training mode:} Standard ($\mathcal{L}_{\text{cls}}$ only) vs.\ Attention-guided ($\mathcal{L}_{\text{cls}} + \lambda \cdot \mathcal{L}_{\text{attn}}$)
\end{itemize}

This yields four conditions per backbone. Four CNN backbones receive all four conditions ($4 \times 4 = 16$ configurations); Swin-V2-T receives only the RealData/Standard configuration. It is excluded from the attention-guided mode because the activation-alignment loss operates on convolutional feature maps, which transformers do not provide; and from the augmented setting because it already attains perfect classification on RealData (F1 = 1.000), which leaves no headroom to assess a \emph{classification} benefit from additional synthetic data (this does not imply that augmentation could not affect its localization, which is simply not studied here). Swin-V2-T therefore serves as a transformer reference baseline rather than a cell of the $2 \times 2$ factorial analysis. Each configuration is repeated with 5 random seeds, yielding $16 \times 5 + 1 \times 5 = 85$ trained models.

The factorial design decomposes the combined effect of DDPM augmentation and attention guidance into individual and interaction terms, testing whether the two strategies are complementary or redundant.

\subsection{Statistical Protocol}

All classification and localization metrics are reported as mean $\pm$ standard deviation across 5 seeds. For both, paired $t$-tests compare standard vs.\ attention-guided training within each backbone and data setting (paired by seed), and Cohen's $d$ quantifies effect sizes. A significance threshold of $\alpha = 0.05$ is used throughout. Localization is evaluated per seed on that seed's held-out defective images, following the held-out protocol described in Section~\ref{sec:methodology_localization}; because each seed yields a different held-out subset and a different trained model, this also captures the seed-to-seed variability of the localization estimate. Because the analysis spans multiple architectures and configurations at five seeds each, and no correction for multiple comparisons is applied, the per-setting significance results are treated as exploratory rather than confirmatory; the within-setting standard-versus-attention-guided comparisons are regarded as the strongest evidence. This within-setting design is deliberate: the two modes being compared share the same data, class weights, held-out pool, and checkpoint-selection criterion, and differ only in the training loss, so the comparison isolates the effect of attention guidance from the data-setting confounds discussed in Section~\ref{sec:limitations}.

\section{Results}\label{sec:results}

\subsection{Classification Performance}

Table~\ref{tab:classification} presents held-out classification results (evaluated only on the held-out validation images, which are excluded from classifier gradient updates).

\begin{table}[!t]
\caption{Held-out classification performance (F1 and AUROC as mean $\pm$ std across 5 seeds; Recall as mean). Best F1 per backbone in bold. For ConvNeXt-T, standard and attention-guided training yield classification metrics that coincide to three decimal places (the alignment loss produces no discernible change in its classification; see Section~\ref{sec:convnext}), so the two modes are reported jointly as ``Std/AG.''}\label{tab:classification}
\begin{tabular}{@{}llccc@{}}
\toprule
\textbf{Backbone} & \textbf{Config} & \textbf{F1} & \textbf{AUROC} & \textbf{Recall} \\
\midrule
\multirow{4}{*}{EfficientNetB0}
    & Real, Std      & .888$\pm$.072 & .961$\pm$.048 & .911 \\
    & Real, AG       & .878$\pm$.086 & .949$\pm$.066 & .911 \\
    & Aug, Std       & \textbf{.902$\pm$.063} & .975$\pm$.016 & .842 \\
    & Aug, AG        & .886$\pm$.052 & .977$\pm$.012 & .827 \\
\midrule
\multirow{4}{*}{MobileNetV2}
    & Real, Std      & .844$\pm$.067 & .962$\pm$.031 & .867 \\
    & Real, AG       & .878$\pm$.058 & .971$\pm$.018 & .889 \\
    & Aug, Std       & \textbf{.883$\pm$.051} & .983$\pm$.015 & .877 \\
    & Aug, AG        & .880$\pm$.046 & .986$\pm$.013 & .861 \\
\midrule
\multirow{4}{*}{ResNet50}
    & Real, Std      & \textbf{.624$\pm$.071} & .834$\pm$.036 & .489 \\
    & Real, AG       & .546$\pm$.096 & .774$\pm$.027 & .467 \\
    & Aug, Std       & .314$\pm$.261 & .890$\pm$.038 & .217 \\
    & Aug, AG        & .189$\pm$.229 & .844$\pm$.052 & .126 \\
\midrule
\multirow{2}{*}{ConvNeXt-T}
    & Real, Std/AG   & \textbf{.988$\pm$.024} & 1.000$\pm$.000 & .978 \\
    & Aug, Std/AG    & .982$\pm$.036 & 1.000$\pm$.000 & .967 \\
\midrule
Swin-V2-T & Real, Std & \textbf{1.000$\pm$.000} & 1.000$\pm$.000 & 1.000 \\
\bottomrule
\end{tabular}
\end{table}

\begin{figure}[!t]
    \centering
    \includegraphics[width=\textwidth]{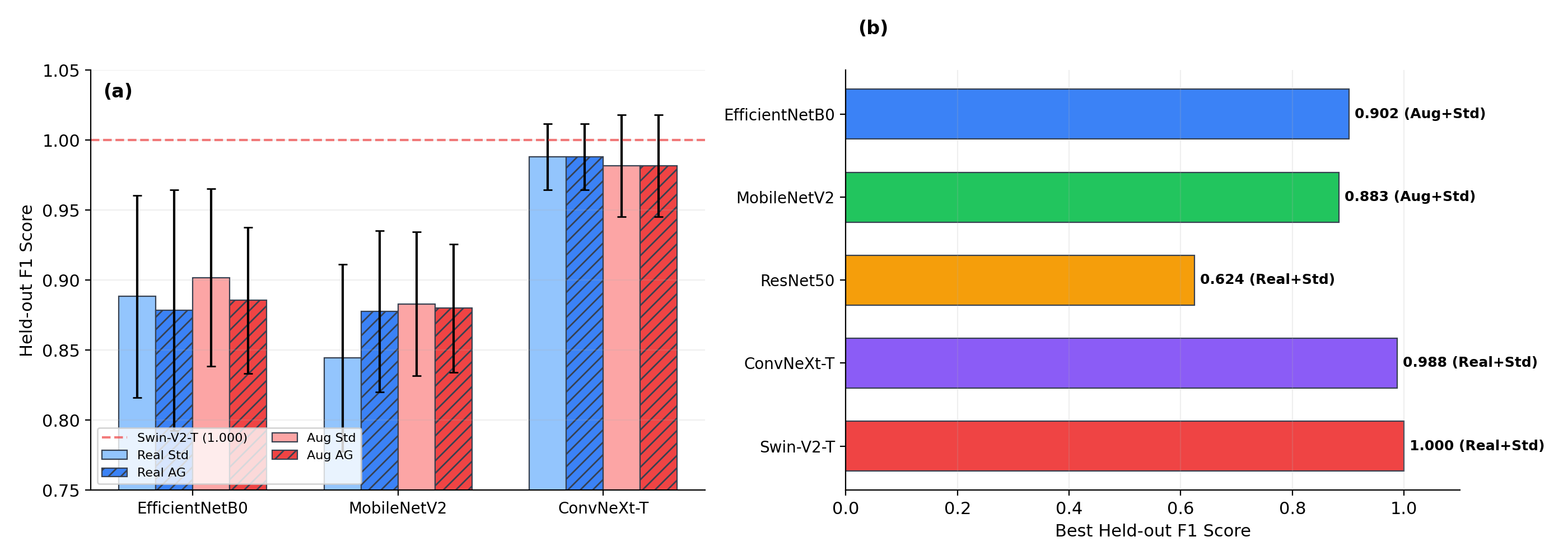}
    \caption{(\textbf{a})~Held-out classification F1 across four training configurations for three CNN architectures. Swin-V2-T reference line at F1=1.000. (\textbf{b})~Best configuration per architecture, showing ConvNeXt-T and Swin-V2-T dominate classification while lighter CNNs achieve 0.88 to 0.90 F1.}
    \label{fig:classification}
\end{figure}

Figure~\ref{fig:classification} visualizes these results. Swin-V2-T achieves perfect classification (F1=1.000). ConvNeXt-T reaches F1=0.988. Among lighter CNNs, EfficientNetB0 and MobileNetV2 achieve F1 in the 0.84 to 0.90 range. ResNet50 performs poorly with LP-FT on this small dataset (F1=0.624 at best), and augmentation destabilizes it further (F1 falling to 0.314 with high variance). This contrasts with the prior study~\cite{rezvani2025enhancing}, in which a fully fine-tuned ResNet50V2 improved with DDPM augmentation (78\% to 93\% accuracy); the difference is consistent with the different variant and the LP-FT protocol used here, under which ResNet50 is the weakest backbone and its localization improvements should be read with that caveat (Section~\ref{sec:limitations}).

Paired $t$-tests across the five seeds confirm that attention-guided training does not significantly change classification for any architecture (all $p > 0.05$); classification performance is broadly comparable between the standard and attention-guided modes. MobileNetV2 shows a slight non-significant improvement on RealData ($+0.034$ F1, $p = 0.202$, Cohen's $d = 0.68$). ConvNeXt-T produces identical standard and attention-guided models ($d = 0.00$), an observation analyzed in Section~\ref{sec:convnext}.

\subsection{Localization: Attention Guidance Effect}

\begin{figure}[!t]
    \centering
    \includegraphics[width=\textwidth]{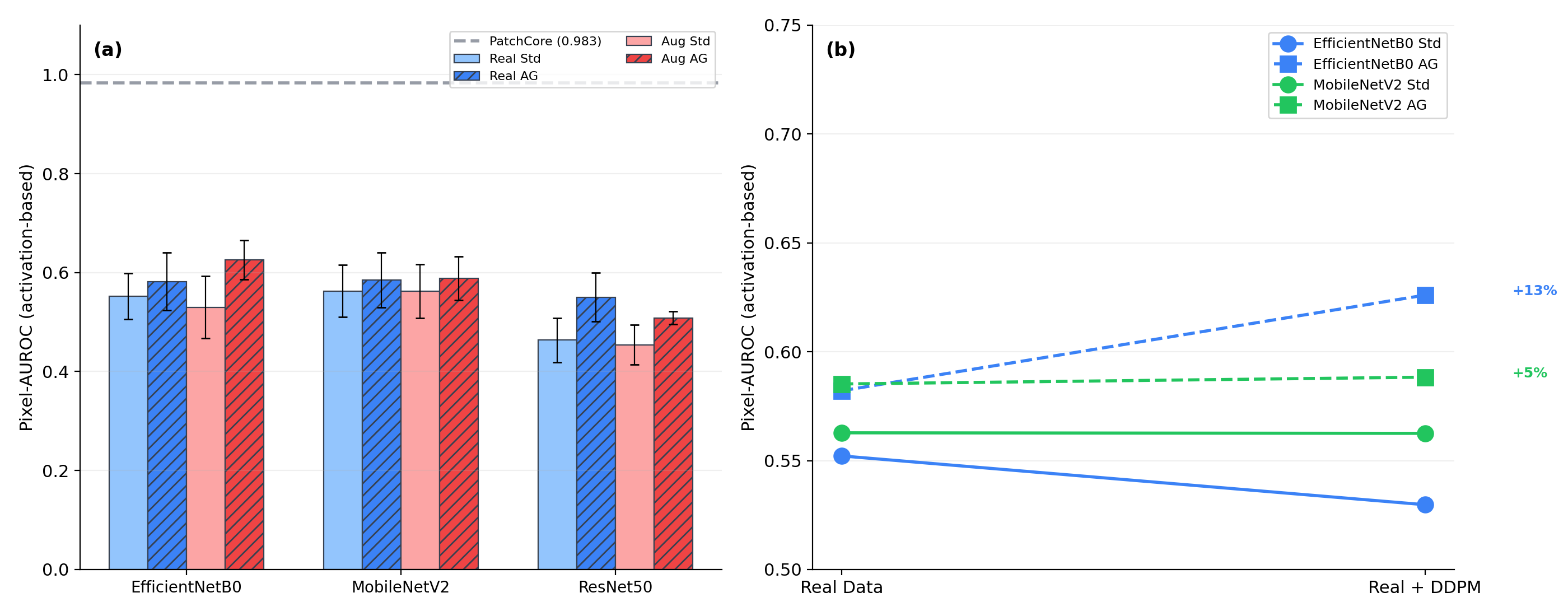}
    \caption{(\textbf{a})~Held-out activation-based localization (Pixel-AUROC): standard vs.\ attention-guided across three CNN architectures and two data settings, with error bars (std across 5 seeds) and the PatchCore reference line. (\textbf{b})~$2 \times 2$ decomposition for EfficientNetB0: DDPM augmentation and attention guidance together show a significant super-additive interaction ($p = 0.002$; Section~\ref{sec:complementarity}).}
    \label{fig:localization}
\end{figure}

Table~\ref{tab:localization} and Figure~\ref{fig:localization} present the key finding: on held-out defect images, attention-guided training improves activation-based localization, with statistically significant gains in four of the eight CNN settings (all four occurring among the three backbones that respond to the loss; ConvNeXt-T, which does not respond, accounts for the two remaining non-significant settings) and large effect sizes.

\begin{table}[!t]
\caption{Held-out localization (activation-based Pixel-AUROC, mean $\pm$ std across 5 seeds). Each model is evaluated only on held-out validation defect images excluded from its gradient updates. $\Delta$ is the relative change from standard (Std) to attention-guided (AG); $p$ is a paired $t$-test (by seed) and $d$ is Cohen's $d$. Significant improvements ($p<0.05$, uncorrected for multiple comparisons) in bold. Swin-V2-T (occlusion) and PatchCore (unsupervised) are localization references.}\label{tab:localization}
\begin{tabular}{@{}llccccr@{}}
\toprule
\textbf{Backbone} & \textbf{Data} & \textbf{Std} & \textbf{AG} & \textbf{$\Delta$} & \textbf{$p$} & \textbf{$d$} \\
\midrule
\multirow{2}{*}{EfficientNetB0}
    & Real & 0.552$\pm$.047 & 0.582$\pm$.058 & +5.4\%  & 0.051 & 1.23 \\
    & +DDPM & 0.531$\pm$.063 & \textbf{0.627$\pm$.040} & \textbf{+18.0\%} & \textbf{0.005} & 2.57 \\
\midrule
\multirow{2}{*}{MobileNetV2}
    & Real & 0.563$\pm$.053 & \textbf{0.585$\pm$.055} & \textbf{+4.0\%}  & \textbf{0.020} & 1.68 \\
    & +DDPM & 0.562$\pm$.054 & 0.588$\pm$.044 & +4.5\%  & 0.238 & 0.62 \\
\midrule
\multirow{2}{*}{ResNet50}
    & Real & 0.464$\pm$.045 & \textbf{0.550$\pm$.049} & \textbf{+18.7\%} & \textbf{0.008} & 2.18 \\
    & +DDPM & 0.454$\pm$.040 & \textbf{0.509$\pm$.013} & \textbf{+12.0\%} & \textbf{0.023} & 1.61 \\
\midrule
\multirow{2}{*}{ConvNeXt-T$^\dagger$}
    & Real & 0.680$\pm$.307 & 0.645$\pm$.305 & $-5.1$\%  & 0.656 & $-0.22$ \\
    & +DDPM & 0.629$\pm$.169 & 0.503$\pm$.012 & $-20.0$\% & 0.217 & $-0.65$ \\
\midrule
Swin-V2-T & Real & \multicolumn{2}{c}{0.705$\pm$.076 (occlusion)} & n/a & n/a & n/a \\
PatchCore & Good-only & \multicolumn{2}{c}{\textbf{0.983} (unsupervised)} & n/a & n/a & n/a \\
\bottomrule
\end{tabular}
\footnotetext{$^\dagger$ ConvNeXt-T shows no consistent effect and very high seed variance; a plausible explanation is that its depthwise-convolution activations are spatially uninformative, which would leave the alignment loss with little useful gradient (Section~\ref{sec:convnext}).}
\end{table}

On held-out images, attention guidance produces statistically significant localization improvements for EfficientNetB0 with augmentation (+18.0\%, $p=0.005$, $d=2.57$), MobileNetV2 on real data (+4.0\%, $p=0.020$), and ResNet50 in both settings (+18.7\%, $p=0.008$ on Real; +12.0\%, $p=0.023$ with DDPM). Effect sizes are large ($d = 1.6$ to $2.6$). EfficientNetB0 on real data shows a positive trend that narrowly misses significance (+5.4\%, $p=0.051$). ConvNeXt-T shows no effect (negative point estimates, non-significant, and very high seed variance), which is analyzed in Section~\ref{sec:convnext}. Notably, ResNet50's standard activation maps score below 0.5 Pixel-AUROC (0.464 and 0.454), i.e., slightly anti-correlated with defect regions, so attention guidance here corrects a mis-localized signal rather than sharpening an already-correct one. Because these results are measured exclusively on images excluded from classifier gradient updates, they indicate generalization of the learned spatial behavior beyond the gradient-updated images rather than memorization of the supervised masks.

\subsection{DDPM and Attention Guidance Appear Complementary}
\label{sec:complementarity}

Figure~\ref{fig:localization}b summarizes the $2 \times 2$ point estimates. For EfficientNetB0, relative to the RealData/Standard baseline (Pixel-AUROC = 0.552):
\begin{itemize}
    \item DDPM alone: 0.552 $\rightarrow$ 0.531 ($-3.8\%$; slightly negative)
    \item Attention alone: 0.552 $\rightarrow$ 0.582 ($+5.4\%$)
    \item \textbf{Combined}: 0.552 $\rightarrow$ \textbf{0.627} ($+13.6\%$)
\end{itemize}

The combined effect ($+13.6\%$) exceeds the sum of the individual effects ($+1.6\%$): neither ingredient does much alone (DDPM augmentation alone is slightly harmful), yet together they yield the largest gain. A two-way (data $\times$ training-mode) interaction test on the per-seed values confirms this for EfficientNetB0: the interaction is positive and significant (mean interaction $+0.066$ Pixel-AUROC, $t(4)=7.0$, $p=0.002$), consistent with a super-additive effect in which DDPM augmentation supplies more defective examples and the attention loss helps the model learn from the correct spatial regions in the enlarged training set. The same test is not significant for MobileNetV2 ($p=0.888$; approximately additive, with DDPM alone neutral at $-0.2\%$ and attention contributing a consistent $\sim$4\% in both settings) or ResNet50 ($p=0.364$; a sub-additive trend in which attention alone at $+18.7\%$ exceeds the combination at $+9.7\%$, consistent with the maskless DDPM images diluting the attention signal). The interaction is therefore architecture-dependent, clearest for EfficientNetB0. Because the RealData and AugmentedData settings also differ in held-out composition (Section~\ref{sec:methodology_localization}) and effective class weighting (Section~\ref{sec:experimental_setup}), and because the augmented cells involve generator exposure (Section~\ref{sec:limitations}), this interaction is interpreted as supportive rather than a clean isolation of a super-additive mechanism; the within-setting standard-versus-attention-guided comparisons remain the primary evidence.

\subsection{Classification and Localization Trade-off}

Among the architectures that respond to activation-based attention supervision, EfficientNetB0 with DDPM augmentation and attention guidance offers the best balance of the two objectives on held-out data (F1 = 0.886, activation-based Pixel-AUROC = 0.627): attention supervision lifts its localization substantially while leaving classification statistically unchanged ($p>0.05$). This qualification is important. ConvNeXt-T attains both higher classification (F1 = 0.98) and higher Grad-CAM localization (0.877), but it does not respond to attention supervision at all (Section~\ref{sec:convnext}); its stronger raw numbers therefore come from the backbone itself rather than from the method studied here. No supervised classifier approaches PatchCore's purpose-built localization performance (0.983), which is expected: PatchCore directly models pixel-level normality, whereas a classifier optimizes an image-level decision and its spatial map is a byproduct. Results are not broken down by defect type, since each seed holds out only a few defective images per category, so per-type estimates are too small to support reliable conclusions.

\subsection{Grad-CAM Comparison}

\begin{figure}[!t]
    \centering
    \includegraphics[width=0.7\textwidth]{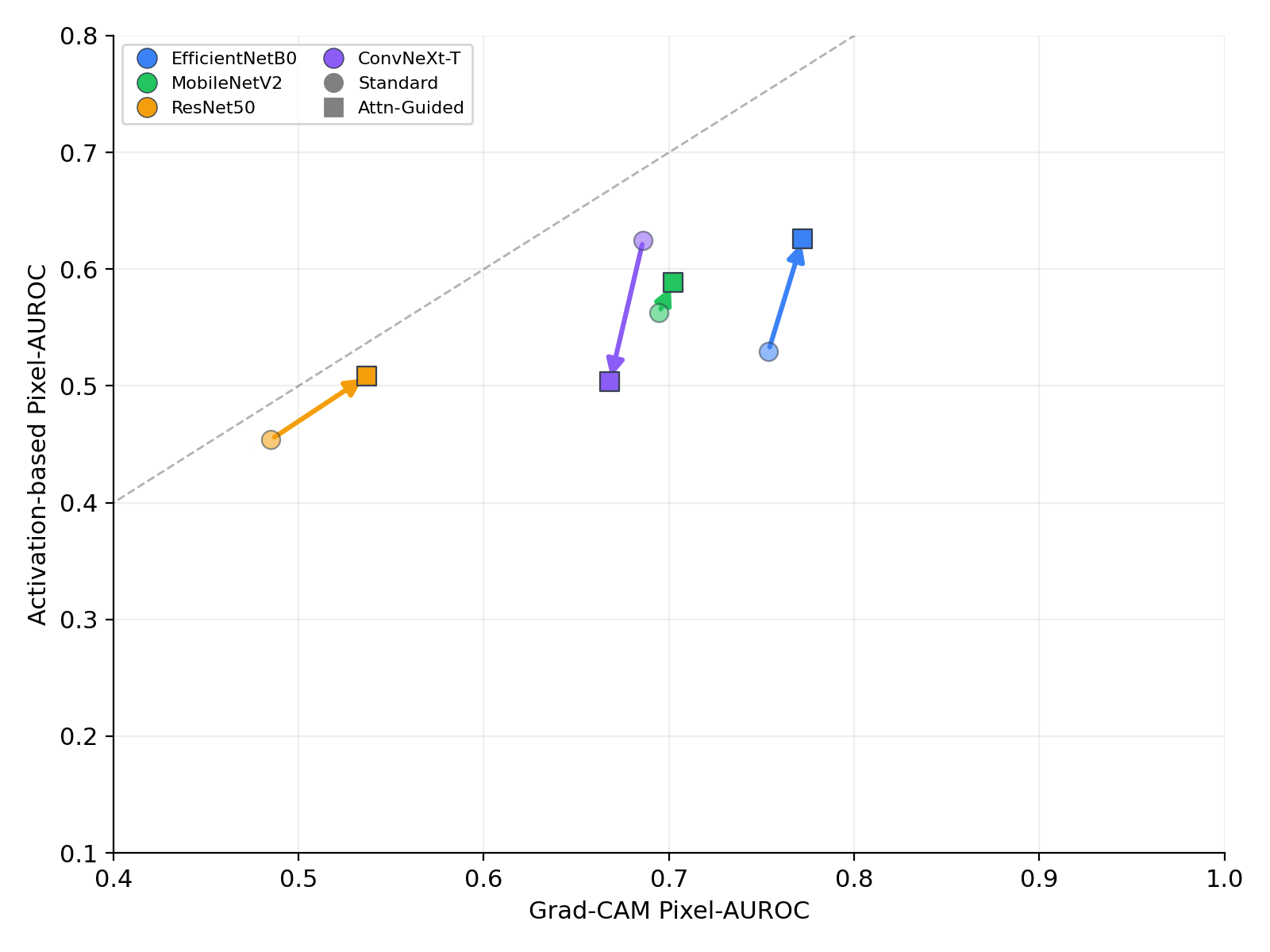}
    \caption{Held-out activation-based vs.\ Grad-CAM localization for the four CNN backbones (AugmentedData-trained, evaluated on held-out defect images). Arrows show the shift from standard (circle) to attention-guided (square) training. All points fall below the diagonal, indicating that Grad-CAM outperforms raw activation maps. Arrows point upward for EfficientNetB0, MobileNetV2, and ResNet50, showing that attention guidance improves activation-based localization while leaving Grad-CAM largely unchanged. ConvNeXt-T's arrow does not point up: its activation map is spatially uninformative and high-variance, and the alignment loss does not improve it, even though its Grad-CAM localization is the strongest of all backbones (see Section~\ref{sec:convnext}).}
    \label{fig:gc_vs_act}
\end{figure}

Figure~\ref{fig:gc_vs_act} compares the two localization evaluation methods on held-out images. Grad-CAM achieves higher Pixel-AUROC than raw activation maps (e.g., EfficientNetB0 with DDPM: 0.772 vs.\ 0.627), because gradient weighting adds class-discriminative information that the channel-mean activation lacks. Attention guidance, which directly optimizes the channel-mean activation, primarily improves the activation-based metric; its transfer to Grad-CAM is smaller but consistent in direction (EfficientNetB0 +DDPM: 0.752$\rightarrow$0.772; ResNet50: 0.485$\rightarrow$0.537 with DDPM, 0.543$\rightarrow$0.579 on real). ConvNeXt-T is informative here: its Grad-CAM localization is the strongest of all CNNs (0.877 on real data) yet is \emph{identical} for standard and attention-guided training, confirming that discriminative spatial information exists in its features but is inaccessible to, and therefore unmodifiable by, the activation-mean alignment loss (Section~\ref{sec:convnext}). Table~\ref{tab:gradcam} reports the full Grad-CAM results. Consistent with the mechanism above, attention guidance produces a statistically significant Grad-CAM improvement only for ResNet50 (+6.6\%, $p=0.044$ on real; +10.7\%, $p=0.024$ with DDPM); for the other backbones the Grad-CAM change is small and non-significant.

\begin{table}[!t]
\caption{Held-out Grad-CAM localization (Pixel-AUROC, mean $\pm$ std across 5 seeds), reported as in Table~\ref{tab:localization}. $\Delta$ is the standard$\rightarrow$attention-guided change; $p$ is a paired $t$-test (by seed) and $d$ is Cohen's $d$. Grad-CAM values exceed the activation-based ones (cf.\ Table~\ref{tab:localization}); attention guidance transfers to Grad-CAM significantly only for ResNet50 (significant improvements in bold).}\label{tab:gradcam}
\begin{tabular}{@{}llccccr@{}}
\toprule
\textbf{Backbone} & \textbf{Data} & \textbf{Std} & \textbf{AG} & \textbf{$\Delta$} & \textbf{$p$} & \textbf{$d$} \\
\midrule
\multirow{2}{*}{EfficientNetB0}
    & Real  & 0.749$\pm$.041 & 0.753$\pm$.041 & +0.5\%  & 0.526 & 0.31 \\
    & +DDPM & 0.752$\pm$.064 & 0.772$\pm$.039 & +2.6\%  & 0.323 & 0.50 \\
\midrule
\multirow{2}{*}{MobileNetV2}
    & Real  & 0.623$\pm$.086 & 0.623$\pm$.063 & 0.0\%   & 0.993 & 0.00 \\
    & +DDPM & 0.696$\pm$.073 & 0.703$\pm$.051 & +1.1\%  & 0.714 & 0.18 \\
\midrule
\multirow{2}{*}{ResNet50}
    & Real  & 0.543$\pm$.023 & \textbf{0.579$\pm$.025} & \textbf{+6.6\%}  & \textbf{0.044} & 1.30 \\
    & +DDPM & 0.485$\pm$.044 & \textbf{0.537$\pm$.026} & \textbf{+10.7\%} & \textbf{0.024} & 1.59 \\
\midrule
\multirow{2}{*}{ConvNeXt-T}
    & Real  & 0.877$\pm$.026 & 0.877$\pm$.026 & $-0.1$\% & 0.937 & $-0.04$ \\
    & +DDPM & 0.680$\pm$.211 & 0.661$\pm$.232 & $-2.8$\% & 0.157 & $-0.78$ \\
\bottomrule
\end{tabular}
\end{table}

\subsection{Qualitative Heatmap Comparison}

\begin{figure}[!t]
    \centering
    \includegraphics[width=\textwidth]{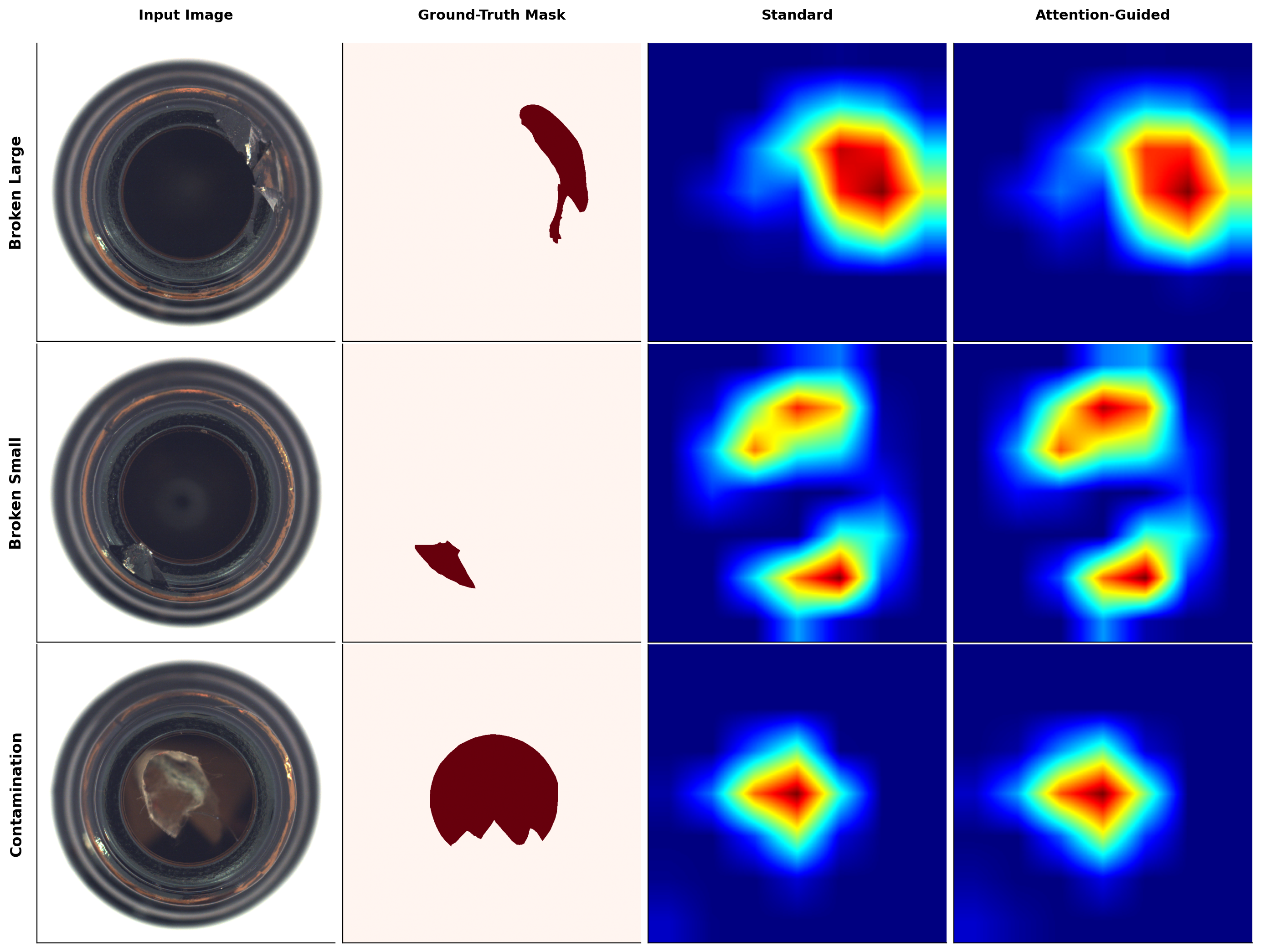}
    \caption{Qualitative comparison of activation heatmaps for EfficientNetB0 (AugmentedData). Columns: input image, ground-truth mask, standard training heatmap, attention-guided heatmap. Attention-guided models produce activations that are visually more concentrated on the actual defect regions.}
    \label{fig:heatmaps}
\end{figure}

Figure~\ref{fig:heatmaps} provides qualitative evidence of the attention guidance effect. For all three defect types, the attention-guided model produces heatmaps that are more spatially concentrated on the defect region compared to the standard model, whose activations are more diffuse across the image.

\section{Discussion}\label{sec:discussion}

\subsection{Why Attention Supervision Works for Some Architectures}
\label{sec:convnext}

The results reveal a clear pattern: architectures with weaker natural spatial representations benefit most from attention supervision. ResNet50, which performs worst at classification on this small dataset (held-out F1 = 0.62), shows the largest and most significant localization improvement (+18.7\%, $p=0.008$). EfficientNetB0 benefits strongly when augmentation supplies additional defective examples (+18.0\%, $p=0.005$). MobileNetV2 shows a smaller but significant gain (+4.0\%, $p=0.020$ on real data).

ConvNeXt-T shows no consistent effect: its classification and Grad-CAM localization are unchanged between the two modes (Grad-CAM Pixel-AUROC of 0.877 for both, to three decimals), while its activation-based estimates differ between modes but are dominated by seed-to-seed noise (std up to $\pm 0.31$ across seeds), so no reliable effect can be established. This does not appear to stem from poor spatial features; ConvNeXt-T in fact achieves the strongest Grad-CAM localization of all CNNs (0.877 on RealData). A more likely explanation lies in the layer the loss operates on. For EfficientNetB0, MobileNetV2, and ResNet50 the hooked last convolution is a $1 \times 1$ pointwise layer that mixes channels, so its channel-mean is spatially informative; for ConvNeXt-T it is a $7 \times 7$ depthwise convolution whose per-channel maps are spatially local and whose channel-mean is largely flat, so the alignment loss appears to receive little useful gradient and leaves the trained model essentially unchanged. This also explains why MobileNetV2, built from depthwise-separable blocks yet exposing a pointwise final convolution, still responds to the loss ($+4.0\%$, $p=0.020$): what matters is the hooked layer, not whether the architecture uses depthwise convolutions elsewhere. Guiding ConvNeXt-T effectively would likely require an architecture-specific attention signal, for example one targeting the depthwise outputs or an attention-style pooling directly. The contrast is instructive: the discriminative spatial information clearly \emph{exists} in its features (Grad-CAM recovers it) yet stays inaccessible to the activation-mean approach, marking a practical boundary on where the method applies.

This suggests that attention supervision is most valuable in resource-constrained scenarios where lightweight models are deployed but spatial precision is important, precisely the conditions encountered in real-time manufacturing inspection.

\subsection{Mixed Supervision and Practical Deployment}

In the AugmentedData configuration, only 63 of 332 training images (19\%) have masks. This mirrors manufacturing practice, where exhaustive pixel-level annotation is rarely feasible. The method gracefully degrades: samples without masks receive classification loss only, while the available masks provide spatial guidance at no additional labeling cost beyond what already exists for evaluation.

\subsection{Relationship to Prior Work}

The proposed approach builds directly on~\cite{rezvani2025enhancing}, which demonstrated that DDPM-generated synthetic images improve classification accuracy. This paper extends that finding by showing that the same augmentation pipeline, when combined with attention-guided training, additionally improves spatial localization. In quantitative terms,~\cite{rezvani2025enhancing} reported only an image-level gain (ResNet50V2 accuracy from 78\% to 93\%) and did not evaluate localization; the present work supplies that missing dimension, improving held-out activation-based Pixel-AUROC by up to +18.7\%, while still trailing the unsupervised PatchCore reference (0.983) that marks the ceiling of a purpose-built localizer. The two contributions address independent problems (data scarcity and spatial precision) and are complementary for lightweight architectures (Section~\ref{sec:convnext}).

Compared to Ross et al.~\cite{ross2017right} (``Right for the Right Reasons''), the approach here differs in three ways: (1)~it supervises activation maps rather than input gradients, avoiding unstable double-backward passes; (2)~it operates in a mixed-supervision setting where only a subset of images have masks; (3)~it targets industrial defect detection rather than sentiment analysis.

\subsection{PatchCore as a Localization Reference}

PatchCore achieves Pixel-AUROC of 0.983, far exceeding all supervised methods. This is expected: PatchCore directly models normality and detects deviations at the patch level, while supervised classifiers optimize for class boundaries rather than spatial precision. However, PatchCore requires storing a feature memory bank and cannot produce class-discriminative explanations. The approaches serve complementary roles: PatchCore for precise localization, supervised models for efficient deployment with optional attention guidance.

\subsection{Limitations}
\label{sec:limitations}

\begin{enumerate}
    \item \textbf{Single dataset.} Results are on MVTec-AD bottle only. Generalization to other categories requires validation.
    \item \textbf{Cross-setting comparisons are indicative.} Two factors differ between the RealData and AugmentedData settings. First, the inherited class-weight formula up-weights the defective class $\approx 1.9\times$ in RealData but only $\approx 0.97\times$ in AugmentedData, where the synthetic defectives enlarge the class (Section~\ref{sec:experimental_setup}). Second, the held-out composition differs between the two settings (Section~\ref{sec:methodology_localization}). The within-setting attention-guidance comparisons share identical weights and held-out pools and are therefore unaffected; the cross-setting decomposition (the super-additive interaction of Section~\ref{sec:complementarity}), although statistically significant for EfficientNetB0, inherits both differences and so is read as supportive rather than as an isolated interaction mechanism. The within-setting results remain the primary evidence for the method.
    \item \textbf{Held-out set doubles as a validation set.} The held-out images used for localization and classification are the per-seed 15\% validation split, which is also used for early stopping and best-checkpoint selection (Section~\ref{sec:methodology_localization}). They are excluded from gradient updates and mask supervision, but they are not a fully untouched test set; reported metrics should be interpreted as held-out validation performance. Because checkpoints are selected by validation classification loss rather than by any localization metric, and because the identical protocol is applied to both training modes, this does not bias the within-setting localization comparison that constitutes the paper's primary evidence.
    \item \textbf{Indirect exposure through augmentation.} The 60 DDPM-generated images were produced in prior work~\cite{rezvani2025enhancing} by an unconditional generator trained on the real defective set, before the per-seed held-out protocol used here was defined. The augmented-data setting therefore involves indirect exposure to defectives that are later held out for evaluation, so those results are not interpreted as strictly leakage-free. This does not affect the paper's primary evidence: the RealData comparisons use no synthetic data and are unaffected, and within the augmented setting the exposure is identical for the standard and attention-guided modes, so it cannot account for the difference between them.
    \item \textbf{Multiple comparisons.} Significance is assessed per architecture and configuration at five seeds without correction for multiple comparisons. The per-setting significance counts are therefore exploratory; the within-setting standard-versus-attention-guided comparisons are the primary evidence.
    \item \textbf{Small held-out localization set.} MVTec-AD bottle contains just 63 defective images, so each seed holds out only 9 to 12. This is mitigated by aggregating across five seeds (45 to 57 held-out evaluations per configuration) and reporting variance, though per-defect-type breakdowns remain too sparse to analyze, and larger benchmarks would yield tighter estimates.
    \item \textbf{ConvNeXt-T insensitivity.} ConvNeXt-T's depthwise convolutions yield spatially uninformative channel-mean activations, which appears to render the attention alignment loss ineffective despite strong Grad-CAM localization (0.877 pAUROC). An architecture-specific attention signal would be needed; gradient-flow diagnostics to confirm the mechanism were not pursued, so this remains a plausible rather than a verified explanation.
    \item \textbf{ResNet50 performance.} ResNet50 performs poorly with LP-FT on this small dataset, limiting the interpretability of its localization improvements.
    \item \textbf{No mask-conditioned augmentation.} The DDPM is unconditional. Mask-conditioned diffusion~\cite{defectfill2025} could produce synthetic images with paired masks, enabling attention supervision on generated samples as well.
\end{enumerate}

\section{Conclusions}\label{sec:conclusions}

This paper studied a lightweight activation-alignment approach that uses ground-truth defect masks as an auxiliary spatial-supervision signal for industrial defect classification networks, in the specific setting of partial-mask supervision combined with maskless DDPM-generated defective samples within a mixed-supervision framework. Across 85 trained models (four CNN backbones under a $2 \times 2$ design across five seeds, plus a Swin-V2-T reference baseline), the results show that:

\begin{enumerate}
    \item \textbf{Attention supervision improves localization on held-out data.} Measured only on held-out validation defect images excluded from gradient updates, activation-based Pixel-AUROC improves by up to +18.7\% (ResNet50, $p=0.008$) and +18.0\% (EfficientNetB0 with augmentation, $p=0.005$), with gains reaching significance in four of eight CNN settings (all four among the three responsive backbones; Cohen's $d=1.6$--$2.6$; uncorrected for multiple comparisons) and no statistically significant change in classification performance.
    \item \textbf{DDPM augmentation and attention supervision are complementary for EfficientNetB0.} The combined effect (+13.6\% over the baseline) exceeds the sum of the individual effects (+1.6\%), and a data $\times$ training-mode interaction test is significant ($p=0.002$), consistent with a super-additive effect in which neither ingredient does much alone. The interaction is architecture-dependent (no significant interaction for MobileNetV2, a non-significant sub-additive trend for ResNet50) and, being a cross-setting comparison, is read as supportive of the within-setting results rather than as their replacement.
    \item \textbf{Architectures with weaker spatial representations benefit most.} The method is most valuable for lightweight models deployed in resource-constrained manufacturing environments.
    \item \textbf{Mixed supervision is practical.} Only 19\% of training images need masks. The method gracefully handles maskless samples (DDPM-generated images receive classification loss only).
\end{enumerate}

Future work should explore mask-conditioned diffusion models that generate synthetic image/mask pairs, enabling attention supervision on all training samples. Architecture-specific attention extraction for modern designs (ConvNeXt, transformers) and extension to multi-class defect classification are additional directions.

\backmatter

\bmhead{Author contributions}
Conceptualization, S.R.B.; methodology, S.R.B. and M.S.; software, S.R.B. and G.T.M.; validation, S.R.B., H.A. and M.S.; formal analysis, S.R.B. and M.S.; investigation, S.R.B., M.S. and G.T.M.; data curation, S.R.B. and G.T.M.; writing---original draft preparation, S.R.B.; writing---review and editing, H.A., T.B., M.S. and G.T.M.; visualization, S.R.B. and M.S.; supervision, T.B. and H.A. All authors have read and agreed to the published version of the manuscript.

\bmhead{Funding}
This research received no external funding.

\bmhead{Data availability}
The MVTec Anomaly Detection Dataset is publicly available under CC BY-NC-SA 4.0 at \url{https://www.mvtec.com/company/research/datasets/mvtec-ad}. Code is available at \url{https://github.com/Actual-Reality/Glass-Defect-Detection-Attention-Supervision}.

\bmhead{Competing interests}
The authors declare no competing interests.


\end{document}